\documentclass[preprint,12pt]{elsarticle}

\usepackage{amssymb}
\usepackage{amsmath}
\usepackage{graphicx}
\usepackage{multirow}%
\usepackage{amsfonts}%
\usepackage{amsthm}%
\usepackage{mathrsfs}%
\usepackage[title]{appendix}%
\usepackage{xcolor}%
\usepackage{textcomp}%
\usepackage{manyfoot}%
\usepackage{booktabs}%
\usepackage{float}
\usepackage{subfig}
\usepackage{subfloat}
\usepackage{algorithm}%
\usepackage{algorithmicx}%
\usepackage{algpseudocode}%
\usepackage{listings}%
\usepackage{makecell}%
\usepackage[colorlinks,
            linkcolor=red, 
             anchorcolor=blue, 
            citecolor=green, 
            ]{hyperref}

\journal{Arxiv}

\begin{document}

\begin{frontmatter}



\title{Co-Paced Learning Strategy Based on Confidence for the Flying Bird Object Detection Model Training}


\author[inst1]{Zi-Wei Sun}
\author[inst2]{Ze-Xi hua \corref{label1}}
\author[inst2]{Heng-Chao Li}
\author[inst2]{Yan Li}

\affiliation[inst1]{organization={School of Computer Engineering, Chengdu Technological University},
            city={Chengdu},
            postcode={611730}, 
            country={China}}
\affiliation[inst2]{organization={School of Information Science and Technology, Southwest Jiaotong University},
            city={Chengdu},
            postcode={611756}, 
            country={China}}
\cortext[label1]{Corresponding author at: School of Information Science and Technology, Southwest Jiaotong University, Chengdu 611756, China. Email address: xx\_zxhua@swjtu.edu.cn}

\begin{abstract}
The flying bird objects captured by surveillance cameras exhibit varying levels of recognition difficulty due to factors such as their varying sizes or degrees of similarity to the background. To alleviate the negative impact of hard samples on training the Flying Bird Object Detection (FBOD) model for surveillance videos, we propose the Co-Paced Learning strategy Based on Confidence (CPL-BC) and apply it to the training process of the FBOD model. This strategy involves maintaining two models with identical structures but different initial parameter configurations that collaborate with each other to select easy samples for training, where the prediction confidence exceeds a set threshold. As training progresses, the strategy gradually lowers the threshold, thereby gradually enhancing the model's ability to recognize objects, from easier to more hard ones. Prior to applying CPL-BC, we pre-trained the two FBOD models to equip them with the capability to assess the difficulty of flying bird object samples. Experimental results on two different datasets of flying bird objects in surveillance videos demonstrate that, compared to other model learning strategies, CPL-BC significantly improves detection accuracy, thereby verifying the method's effectiveness and advancement.
\end{abstract}


\end{frontmatter}


\section{Introduction}\label{intro}

The detection of flying bird objects holds significant application value across various fields \cite{a_new_skeleton_based_flying_bird_detection, yolov5s_farm_bird_detction, automated_monitoring_for_birds}. Currently, the radar-based method for detecting flying bird objects is widely used \cite{2016_Hoffmann_multistatic_radar, 2017_Jahangirstaring_radar}, but radar equipment suffers from disadvantages such as large size, high cost, and limited visual effectiveness. With the advancement of computer vision, deep learning, and other technologies, there is a growing amount of research focused on using cameras to detect various objects \cite{object_survey}.

\begin{figure*}[!ht]
\centering
\includegraphics[width=5.4in]{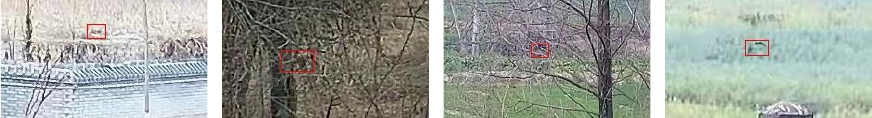}
\caption{Examples of hard samples of flying bird objects (red boxes indicate flying bird objects)}
\label{examples_of_hard_samples}
\end{figure*}

We are working on how to use surveillance cameras to detect flying bird objects \cite{fbod-bmi, fbod-sv, spl_bc_conference, spl_bc}. However, the flying bird objects captured by surveillance cameras pose different levels of identification difficulty due to their varying sizes or degrees of similarity to the background. During model training, the model may be adversely affected by the noise inherent in hard samples (e.g., those illustrated in Figure \ref{examples_of_hard_samples}), thereby reducing detection accuracy \cite{spl_bc}.

To address the aforementioned problem, an intuitive approach would be to train the model using only easy samples and exclude the hard ones (similar to filtering out noisy samples when dealing with data labeled with noise \cite{lu_mentornet, malach_when_to_update, Han_co_teacher, shen_learning_with_bad_training_data}). By doing so, it can, to a certain extent, avoid the interference of ambiguous, hard samples during the model training process, thereby reducing the likelihood of false detections. However, excluding hard samples from the training process may result in poor performance when the model encounters these hard samples. To resolve this contradiction, a new method called the Self-Paced Learning strategy with Easy Sample Prior Based on Confidence (SPL-ESP-BC) was proposed in the literature \cite{spl_bc}. The core idea of this strategy is to initiate learning from easy samples and gradually transition to hard samples, improving the model's recognition ability of the object step by step and effectively mitigating the adverse impact of hard samples on model training. Compared to the loss-based Self-Paced Learning (SPL) strategy \cite{kumar_SPL_for_latent_variable, jiang_easy_samples_first_sp_regularizers, spl_diversity, spl_matrix_factor, spl_cl, evolution_spl}, SPL-ESP-BC \cite{spl_bc} utilizes the prediction confidence of samples to assess their difficulty level. In the context of One-Category object detection tasks (where flying bird object detection falls), using confidence as a criterion offers greater advantages than using loss. However, it should be noted that during the SPL process, bias accumulation may arise in sample selection. Specifically, the model may mistakenly categorize some hard samples as easy ones or vice versa, leading to selection bias. Since the SPL strategy relies on the model itself to evaluate the sample difficulty, the sample selection bias may gradually accumulate during the training process and ultimately affect the model's performance.

In response to the challenge of noisy labeling, literature \cite{Han_co_teaching} proposes a novel and effective learning paradigm called ``Co-teaching". This paradigm involves training two deep neural networks simultaneously. After each iteration completes its forward inference, each network selects a batch of data with sufficiently small losses based on its calculated loss value, considering these data to be more accurate (as they exhibit less loss when processed through the network, indicating that the network's predictions for these data are more consistent with the given labels). These batches of data, deemed ``clean", are then passed to the peer network for parameter updates. Since deep networks have a tendency to prioritize learning from ``clean" data and develop robustness \cite{closer_look_deep_networks}, when data containing noise is mistakenly identified as ``clean" and passed to a peer network, the peer network can, to a certain extent, filter out these noisy data due to its robustness, thereby effectively preventing the accumulation of such selection bias. Through this mutual screening and transmission mechanism, the two networks collaborate to reduce the errors introduced by noisy labels.

Inspired by Co-teaching and combined with the principle of Self-Paced Learning strategy Based on Confidence (SPL-BC), this paper proposes a Co-Paced Learning strategy Based on Confidence (CPL-BC) to optimize the training of the Flying Bird Object Detection (FBOD) model for surveillance video. The core of this strategy is maintaining two object detection models with the same structure but different initial parameter configurations. In each training iteration, the models independently evaluate the confidence of the sample set and select those whose confidence exceeds the currently set threshold as ``easy samples". These selected easy samples are then exchanged between the models for parameter updates. As the training progresses and the number of rounds increases, the confidence threshold is gradually lowered to include more samples, gradually transitioning from easy to hard. This process starts with easy samples and progressively incorporates harder ones, enhancing the model's ability to recognize object features and mitigating the adverse effects of hard samples on training. By alternately assessing sample difficulty using two models, this strategy can, to some extent, prevent the accumulation of sample selection bias.

Before applying the CPL-BC strategy to train the FBOD model, we first independently and randomly initialize two FBOD models with the same structure to ensure they have distinct initial parameter configurations. Then, both models undergo preliminary training to equip them with the ability to assess the difficulty of samples. Subsequently, the CPL-BC strategy is applied during their further training.

The main contributions of this paper are as follows.

\begin{enumerate}
\item{The CPL-BC training strategy for the One-Category object detection model is proposed by integrating the concept of Co-teaching into the SPL-BC strategy. It involves simultaneously training two neural network models, each independently evaluating sample confidence in each iteration. They select ``easy samples" with confidence exceeding the current threshold and pass them to the peer network for updates. As training progresses and rounds increase, the threshold gradually lowers, allowing more samples to train. This avoids selection bias from single-model evaluation.}
\item{The CPL-BC strategy is applied to the training process of the FBOD model. Firstly, two FBOD models with identical structures are independently and randomly initialized to ensure they possess different initial parameter configurations. Subsequently, these two models undergo preliminary training to equip them with the capability to assess the difficulty levels of samples. Then, in the subsequent training phase, the CPL-BC strategy is adopted to further enhance the detection performance of the FBOD model.}
\end{enumerate}

The rest of this paper is structured as follows: Section \ref{Preliminaries} introduces the work related to this paper. Section \ref{The_CPL-BC} describes the CPL-BC in detail. Section \ref{CPL-BC_for_FBOD} introduces how to apply the proposed CPL-BC strategy to the training process of the FBOD model. Section \ref{Experiments} carries out quantitative and qualitative experiments on the proposed method. Section \ref{Conclusion} summarizes the work of this paper.

\section{Preliminaries}\label{Preliminaries}

The main focus of this paper is to innovatively combine the Co-teaching method with the SPL-BC model training strategy, and apply this integrated strategy to the training process of the FBOD model for surveillance video. To pave the way for the subsequent content, in this section, we briefly review the SPL-BC method and introduce the FBOD-SV method \cite{fbod-sv}, which is specifically designed for flying bird object detection in surveillance videos.

\subsection{The SPL-BC Algorithm}

Given a training dataset $\textbf{D} = \left\{ \left (\textbf{x}_i,\text{y}_i \right ) \right \}_{(i=1)}^n$, where $\textbf{x}_i$ and $\text{y}_i$ denote the $i^\text{th}$ input sample and its label, respectively. When the input is $\textbf{x}_i$, $f\left (\textbf{x}_i;\textbf{w} \right )$ represents the prediction result of the model $f$, whose parameter is $\textbf{w}$. Then, the expression of the SPL \cite{kumar_SPL_for_latent_variable} is as follows,
\begin{align}\label{eq:spl}
\min\limits_{\textbf{w}, \textbf{v}} \text{E} \left( \textbf{w}, \textbf{v}, \lambda \right) = \sum\limits_{i=1}^n \left( v_i \text{L} \left (\text{y}_i,f\left (\textbf{x}_i;\textbf{w} \right ) \right ) + g\left( v_i, \lambda \right) \right),
\end{align}
where $\lambda$ is the age parameter that controls the learning speed. $\text{L} \left (\text{y}_i,f\left (\textbf{x}_i;\textbf{w} \right ) \right )$ represents the loss between the predicted result and the corresponding label. $v_i \in \textbf{v}$ $\left( \textbf{v}=[v_1, v_2, \cdots, v_n] \right)$ is the sample weight, weighted to the loss of the $i^{th}$ sample, which is used to control whether the sample participates in training or the degree of participation. $g\left( v_i, \lambda \right)$ is the Self-Paced Regularizer, and the determination of the sample weight depends on it.

The process of SPL (to solve Eq. \eqref{eq:spl}) is as follows: First, determine the optimal weight $\textbf{v}$ of the samples,
\begin{align}\label{eq:splv}
\min\limits_{\textbf{v}} \sum\limits_{i=1}^n \left( v_i \text{L} \left (\text{y}_i,f\left (\textbf{x}_i;\textbf{w} \right ) \right ) + g\left( v_i, \lambda \right) \right),
\end{align}
where the model parameter $\textbf{w}$ is fixed. When the optimal weight $\textbf{v}$ of the samples is determined, it is fixed and weighted to the sample loss. The optimal weight $\textbf{w}$ of the model is obtained by optimizing the following Weighted Loss Function,
\begin{align}\label{eq:splw}
\min\limits_{\textbf{w}} \sum\limits_{i=1}^n  v_i \text{L} \left (\text{y}_i,f\left (\textbf{x}_i;\textbf{w} \right ) \right ).
\end{align}
Then, taking a particular strategy, increase the age parameter $\lambda$ and repeat the process. With the increase of the age parameter, fixing one of $\textbf{w}$ and $\textbf{v}$ alternately to optimize the other, the samples to be learned can be gradually included in the training from easy to hard, thus realizing the SPL process.

When seeking the optimal sample weight parameter $\textbf{v}$, since the model weight $\textbf{w}$ is fixed and the loss of the $i^{th}$ sample is a constant, the optimal value of the weight $v_i$ is uniquely determined by the corresponding Minimizer Function $\sigma\left( \lambda, \text{L} \left (\text{y}_i,f\left (\textbf{x}_i;\textbf{w} \right ) \right ) \right)$, and has
\begin{align}\label{eq:spl_minimizer}
\sigma\left( \lambda, l_i \right) l_i + g\left( \sigma\left( \lambda, l_i \right), \lambda \right) \leq v_i l_i + g\left( v_i, \lambda \right), \forall v_i \in [0, 1],
\end{align}
where $l_i = \text{L} \left (\text{y}_i,f\left (\textbf{x}_i;\textbf{w} \right ) \right )$.

If $g\left( v_i, \lambda \right)$ has a concrete analytical form, it is called a Self-Paced Explicit Regularizer. The Minimizer Function can be derived from Eq. \eqref{eq:spl}, which provides a Self-Paced Explicit Regularizer. Table \ref{tab:Minimizer_Functions} shows some classical Self-Paced Explicit Regularizers and their Minimizer Functions (closed-form solutions for the optimal sample weights).
\begin{table}[!ht]
\caption{Some classical Self-Paced Explicit Regularizers and their Minimizer Functions.\label{tab:Minimizer_Functions}}
\centering
\begin{tabular}{c|c|c}
\hline
Names & Regularizers & Minimizer Functions \\
\hline
Hard \cite{kumar_SPL_for_latent_variable} & $-\lambda \sum\limits_{i=1}^n v_i$ & $\begin{cases} 1, &l_i < \lambda,\\0, &{\text { Otherwise}},\end{cases}$ \\
\hline
Linear \cite{jiang_easy_samples_first_sp_regularizers} & $\frac{1}{2} \lambda \sum\limits_{i=1}^n\left( v_i^2 - 2 v_i\right) $ & $\begin{cases} 1 - l_i/\lambda, &l_i < \lambda,\\0, &{\text { Otherwise}},\end{cases}$ \\
\hline
Logarithmic \cite{jiang_easy_samples_first_sp_regularizers} & $\makecell[c]{\sum\limits_{i=1}^n\left( \zeta v_i - \frac{\zeta ^{v_i}}{\text{log} \zeta}\right)\\\zeta = 1 - \lambda, 0 < \lambda < 1}$ & $\begin{cases} \frac{\text{log}\left( l_i + \zeta \right)}{\text{log} \zeta}, &l_i < \lambda,\\0, &{\text { Otherwise}},\end{cases}$ \\
\hline
Polynomial \cite{Gong_Polynomial_regularization} & $\makecell[c]{\lambda \left(\frac{1}{t}\sum\limits_{i=1}^N v_i^t-\sum\limits_{i=1}^N v_i\right) \\ \lambda > 0, t > 1}$ & $\begin{cases} \left(1 - \frac{l_i}{\lambda}\right)^\frac{1}{t-1}, &l_i < \lambda,\\0, &{\text { Otherwise}},\end{cases}$ \\
\hline
\end{tabular}
\end{table}

Based on the convex conjugate theory, Fan et al. \cite{yanbo_SPL_implicit_regularization} prove that during optimization, the optimal sample weight $\textbf{v}^*$ is determined by the Minimizer Function $\sigma (\lambda,l)$, while the analytic form of the Self-Paced Regularizer $\psi \left( v_i, \lambda \right)$ can be unknown. The Self-Paced Regularizer at this time is called the Self-Paced Implicit Regularizer by them. Fan et al. \cite{yanbo_SPL_implicit_regularization} also proposed a Self-Paced Learning framework based on Self-Paced Implicit Regularizers,
\begin{align}\label{eq:spl_implicit}
\min\limits_{\textbf{w}, \textbf{v}} \text{E} \left( \textbf{w}, \textbf{v}, \lambda \right) = \sum\limits_{i=1}^n \left( v_i \text{L} \left (\text{y}_i,f\left (\textbf{x}_i;\textbf{w} \right ) \right ) + \psi \left( v_i, \lambda \right) \right),
\end{align}
where $\psi \left( v_i, \lambda \right)$ is a Self-Paced Implicit Regularizer. Eq. \eqref{eq:spl_implicit} can be solved by selective optimization algorithm. Unlike the ordinary SPL method, the form of the Self-Paced Implicit Regularizer in Eq. \eqref{eq:spl_implicit} can be unknown, and the optimal sample weight $v_i^*$ is determined by the Minimizer Function $\sigma (\lambda,l_i)$ as follows,
\begin{align}\label{eq:spl_mf}
 v_i^* = \sigma\left( \lambda, \text{L} \left (\text{y}_i,f\left (\textbf{x}_i;\textbf{w} \right ) \right ) \right),
\end{align}
where $v_i^*$ represents the optimal weight of the sample $i$. As can be seen from Eq. \eqref{eq:spl_mf}, the difficulty of judging samples by SPL is based on the loss value of the sample, and the judging standard is determined by the SPL age parameter $\lambda$. In the SPL-BC strategy \cite{spl_bc}, this evaluation basis is replaced by the prediction confidence of the sample. Accordingly, the Minimizer Function is also modified, and the modified Minimizer Function is as follows,
\begin{align}\label{eq:mf_bc}
v_i^* = \sigma' \left( \xi, \text{Conf}_{\text{pred}}\left(i\right) \right),
\end{align}
where $\text{Conf}_{\text{pred}}\left(i\right)$ represents the prediction confidence of the $i^{th}$ input sample, $\xi$ is a parameter inversely related to the age parameter $\lambda$, and $\sigma'$ represents the Minimizer Function based on confidence.

With the fixed sample weights, Eq. \eqref{eq:splw} optimizes the model parameters. Then, the optimized model parameters are fixed, and Eq. \eqref{eq:mf_bc} optimizes the sample weight. After that, adjust the parameter $\xi$. This process is iterated alternately until a set number of times is reached. This iterative optimization process constitutes the Self-Paced Learning strategy Based on Confidence (SPL-BC).

In the process of SPL-BC, the parameter $\xi$ gradually decreases as the number of training iterations increases. This implies that samples initially considered hard, due to their lower prediction confidence, will become more likely to be included in the training process as the threshold $\xi$ decreases. Specifically, $\xi$ can be viewed as a dynamic threshold parameter closely related to the prediction confidence of samples, determining which hard samples (i.e., samples with lower prediction confidence) are included in the current training round. This adjustment strategy, where $\xi$ gradually decreases with the increasing number of training iterations, is referred to as the Training Schedule strategy.

\subsection{The FBOD-SV Method}

The FBOD-SV \cite{fbod-sv} is designed to detect flying birds in surveillance videos. Its input comprises consecutive frames of video images, and its output is the position information of flying birds on these images, specifically represented as bounding boxes. The implementation workflow of FBOD-SV is as follows: Firstly, a processing unit focused on fusing multi-frame video image information, known as the Correlation Attention Feature Aggregation unit (Co-Attention-FA), is utilized to integrate this information, with special attention given to the flying bird objects in the video. Subsequently, a network structure that first down-samples and then up-samples is employed to extract features of the flying bird objects. These features are then passed to a large feature map for subsequent prediction of flying bird objects. Finally, the model outputs two branches: one branch is responsible for predicting whether feature points (i.e., anchor points) in the feature map belong to a flying bird object and providing the corresponding confidence score (the confidence prediction branch); the other branch is responsible for providing the specific location information of the flying bird object relative to the feature points, i.e., the regression results of the bounding box (the bounding box regression branch).

The FBOD-SV trains the FBOD model using a multi-task loss function, which consists of confidence loss and location regression loss. Specifically, during the training process, the loss value of each anchor point is obtained by calculating the weighted sum of its confidence loss and location regression loss. The formula for the loss function is as follows,
\begin{align}\label{eq:lai}
\text{L} \left( {\text{A}}_{i} \right ) = \text{L}_{\text{Conf}}\left ( {\text{A}}_{i} \right ) + \alpha \text{L}_{\text{Reg}}\left ( {\text{A}}_{i} \right ),
\end{align}
where $\text{A}_{i}$ represents the $i^{th}$ anchor point, $\text{L}_{\text{Conf}}\left ( \cdot \right )$ denotes the confidence loss, which employs the L2 loss. $\text{L}_{\text{Reg}}\left ( \cdot \right )$ represents the location regression loss, utilizing the CIoU loss \cite{zheng_Enhancing_Geometric_Factors_in_Model_Learning_and_Inference}. The $\alpha$ is a balancing parameter for the two types of losses. During training, the total loss is equal to the sum of the losses for all anchor points, formulated as follows,
\begin{align}\label{eq:tl}
\text{L}_\text{Total} &=\frac{1}{\text{N}} \sum{\text{L}\left (\text{A}_i\right)}\notag\\
&=\frac{1}{\text{N}}\left({\sum{\text{L}}_{\text{Conf}}\left (\text{A}_i\right)} + {\alpha}\sum{{\text{L}}_{\text{Reg}}\left (\text{A}_i\right)}\right)\notag\\
&=\frac{1}{\text{N}}\left({\text{L}}_{\text{C}}+{\alpha}{\text{L}}_{\text{R}} \right),
\end{align}
where N is the normalization parameter when the image contains flying bird objects, N is defined as the number of positive anchor point samples involved in training; when the image does not contain flying bird objects, N is defined as a fixed positive number selected based on actual conditions to ensure the stability of the loss function. ${\text{L}}_{\text{C}}$ represents the sum of the confidence losses for all anchor points involved in training, and  ${\text{L}}_{\text{R}}$ represents the sum of the location regression losses for all positive anchor point samples. In the flying bird objects detection method for surveillance videos, anchor points can be classified into negative anchor point samples, positive anchor point samples for flying bird object 1, positive anchor point samples for flying bird object 2, ..., and positive anchor point samples for flying bird object n. Therefore, Eq. \eqref{eq:tl} can also be rewritten as the sum of the losses for each flying bird object sample,
\begin{align}\label{eq:tl_2}
\text{L}_\text{Total} =\frac{1}{\text{N}} \left (\text{L}\left (\text{A}_{neg}\right) + \text{L}\left (\text{A}_{F_1}\right) + ... + \text{L}\left (\text{A}_{F_n}\right)\right),
\end{align}
where $\text{A}_{neg}$ represents the negative anchor sample set, and $\text{A}_{F_i} (i \in (1, ... ,n))$ represents the positive anchor sample set for the flying bird object $i$.


\section{The CPL-BC Method}\label{The_CPL-BC}

The CPL-BC strategy involves maintaining two models that collaboratively select easy samples for each other, where the prediction confidence exceeds a certain threshold, and then training them. As the training progresses, the strategy gradually lowers this threshold to include more samples in the training. Specifically, the CPL-BC strategy comprises the following five steps.

\textbf{Step 1: Initialize Parameters}

Firstly, define a confidence-based Minimizer Function $\sigma'$, which is used to determine the optimal weights for training samples. Next, establish a Training Schedule method $\Psi$ that will be employed to dynamically adjust the criteria for evaluating the easiness or hardness of samples during the training process. Additionally, set an initial confidence threshold-related parameter $\xi$ as the starting point. Subsequently, initialize two models, $f$ and $g$, with identical structures using two different sets of weight parameters,$\textbf{w}_f$ and $\textbf{w}_g$, respectively.

\textbf{Step 2: Obtain the Optimal Weights of the Samples}

Randomly select a batch of data $\{\textbf{X},\textbf{Y}\} \subset \textbf{S}$, input the samples $\textbf{X}$ into model $f$, and obtain the prediction confidence for each sample in this batch of data, as follows,
\begin{align}\label{eq:predict_confidence}
\textbf{C}_f = \text{Conf}\left( f\left (\textbf{X};\textbf{w}_f \right ) \right),
\end{align}
where $\textbf{X}=[\textbf{x}_1,\cdots,\textbf{x}_n ]$, $\textbf{Y}$ is the label of the corresponding sample, $\textbf{S}$ is the training set, $\text{Conf}\left( \cdot \right)$ is a post-processing function to obtain the prediction confidence, and $\textbf{C}_f=[\textbf{c}_{f_1}, \cdots, \textbf{c}_{f_n}]$ represents the prediction confidence of the samples $\textbf{X}$ after inference by model $f$. Then, the prediction confidence $\textbf{C}_f$ is input into the confidence-based Minimizer Function $\sigma'$, resulting in a set of optimal weights $\textbf{V}_g^*$ for this batch of samples, as follows,
\begin{align}\label{eq:predict_confidence}
\textbf{V}_g^* = \sigma'\left(\xi;\textbf{C}_f \right).
\end{align}
Similarly, the samples $\textbf{X}$ are input into model $g$ to obtain the corresponding prediction confidences, and these confidences are input into the confidence-based Minimizer Function $\sigma'$ to get another set of optimal weights $\textbf{V}_f^*$ for this batch of samples.

\textbf{Step 3: Weighted Summation of Sample Losses}

Apply the sample weights $\textbf{V}_g^*$ and $\textbf{V}_f^*$ to the loss functions of model $g$ and model $f$, respectively, for their corresponding samples, which means multiplying the loss of each sample by its respective weight and then summing them up, as follows,
\begin{align}\label{eq:wsl_g}
\textbf{V}_g^* \text{L} \left (\textbf{Y},g\left (\textbf{X};\textbf{w}_g \right ) \right ) = \sum\limits_{i=1}^n v_{g_i}^* \text{L} \left (\text{y}_i,g\left (\textbf{x}_i;\textbf{w}_g \right ) \right ),
\end{align}
\begin{align}\label{eq:wsl_f}
\textbf{V}_f^* \text{L} \left (\textbf{Y},f\left (\textbf{X};\textbf{w}_f \right ) \right ) = \sum\limits_{i=1}^n v_{f_i}^* \text{L} \left (\text{y}_i,f\left (\textbf{x}_i;\textbf{w}_f \right ) \right ).
\end{align}
In this way, samples with larger weights (relatively easy samples) will contribute more to the loss function, facilitating the prioritization of easy samples during the training process.

\textbf{Step 4: Optimize the Model Parameters}

Use the backpropagation algorithm to propagate the loss and employ the Gradient Descent (GD) algorithm to optimize and update the weights $\textbf{w}_g$ of model $g$ and the weights $\textbf{w}_f$ of model $f$.

\textbf{Step 5: Update Training Schedule Parameter}

According to the Training Schedule method determined in Step 1, dynamically adjust the Training Schedule parameter $\xi$ related to the confidence threshold as follows,
\begin{align}\label{eq:wsl_f}
\xi = \Psi \left( \text{EP} \right),
\end{align}
where EP represents the training progress, ranging from [0\%, 100\%]. The update of the Training Schedule parameter gradually lowers the confidence threshold, allowing more hard samples to participate in training in subsequent iterations.

Repeat Steps 2 to 5, gradually increasing the difficulty of the training samples so that they participate more actively in the model training process. Once all training iterations are completed, fix the weight parameters of the models, thereby completing the CPL-BC training process.

\section{Applying the CPL-BC strategy to the training process of the FBOD model}\label{CPL-BC_for_FBOD}

\begin{figure*}[!ht]
\centering
\includegraphics[width=5.4in]{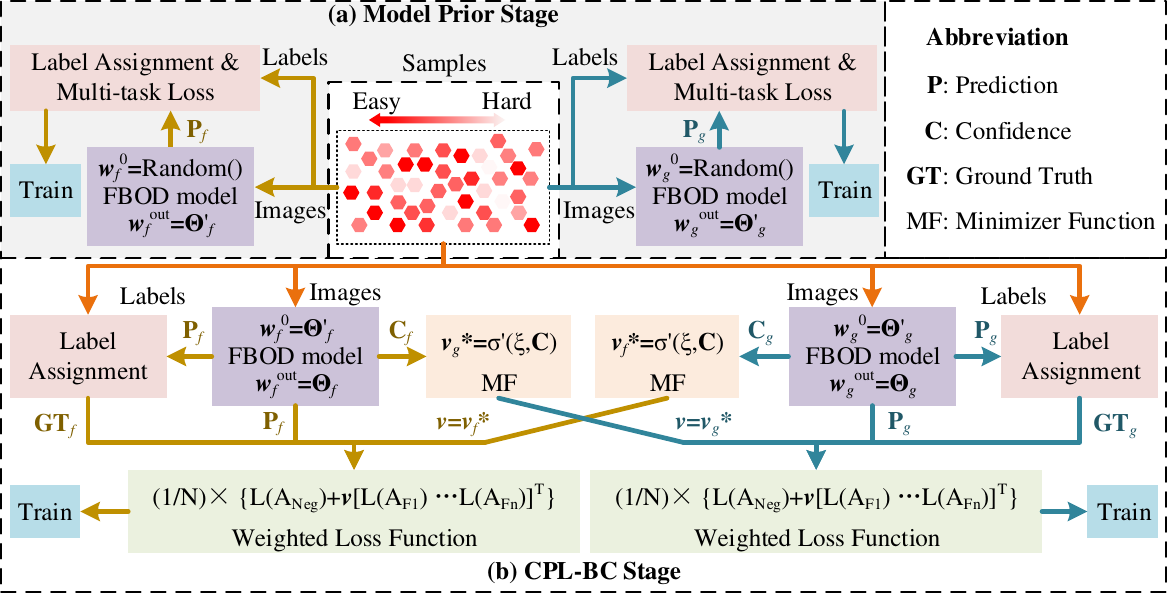}
\caption{Block diagram of the CPL-BC training strategy for the flying bird object detection model for surveillance video.}
\label{fig:CPL-BC_for_FBOD}
\end{figure*}

Figure \ref{fig:CPL-BC_for_FBOD} illustrates the training process of the FBOD model for surveillance video using the proposed CPL-BC strategy. The process includes two main stages: the Model Prior stage and the CPL-BC training stage. In the Model Prior stage, all samples are used to train two FBOD models with randomly initialized weights, enabling them to initially acquire the ability to assess the difficulty level of samples [as shown in Figure \ref{fig:CPL-BC_for_FBOD}(a)]. In the CPL-BC training stage, all samples are reused, and the CPL-BC strategy is applied for further training. At this stage, the models' weights are initialized with the fixed weights obtained after the Model Prior stage [as shown in Figure \ref{fig:CPL-BC_for_FBOD}(b)]. During this stage, the predictive confidence of one model for a sample determines the optimal sample weight for the other model's Weighted Loss Function, controlling whether and to what extent the sample participates in training.

Next, the Weighted Loss Function, the Minimizer Function, and the Training Schedule method involved in applying the CPL-BC strategy to the training of the FBOD model are introduced. Then, the process of applying the CPL-BC strategy to FBOD model training is explained in detail.

\subsection{The Weighted Loss Function}

When applying the CPL-BC strategy to the training process of the FBOD model, the calculated optimal sample weights are directly applied to the loss of each flying bird object sample to obtain the required Weighted Loss Function. By substituting the sample weights into Eq. \eqref{eq:tl_2}, the expression of this Weighted Loss Function can be derived as follows,
\begin{align}\label{eq:wlf_fbod}
\text{L}_\text{Total} &=\frac{1}{\text{N}} \left (\text{L}\left (\text{A}_{neg}\right) + v_i \text{L}\left (\text{A}_{F_1}\right) + ... + v_n \text{L}\left (\text{A}_{F_n}\right)\right)\notag\\
&=\frac{1}{\text{N}} \left (\text{L}\left (\text{A}_{neg}\right) + \boldsymbol{\Vec{v}} \left[ \text{L}\left (\text{A}_{F_1}\right) ...\ \text{L}\left (\text{A}_{F_n}\right)\right]^T \right),
\end{align}
where $\boldsymbol{\Vec{v}}=[v_1\ ...\ v_n]$ is the sample weight corresponding to the loss of each flying bird object sample, which controls which flying bird objects participate in training, or the degree of participation in training, and its value is determined by the Minimizer Function. The form of the Weighted Loss Function is similar to that of applying the SPL-BC strategy to the training process of the FBOD model \cite{spl_bc}. The difference, however, is that the loss weight of each flying bird object sample in the Co-Paced Learning process is determined by another model through inference prediction.

\subsection{The Minimizer Function and the Training Schedule method}

\begin{figure}[!ht]
\centering
\includegraphics[width=3in]{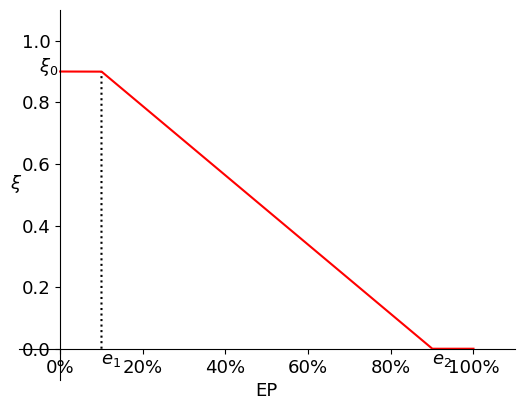}
\caption{The relationship between $\xi$ and training process.}
\label{xi_ep_fig}
\end{figure}

For subsequent experimental comparison, when applying the CPL-BC strategy to the training process of the FBOD model, we adopted the same Minimizer Function and Training Schedule method as those in the literature \cite{spl_bc}. The expression for this Minimizer Function is as follows,
\begin{align}\label{eq:mf_example}
v_i=
\begin{cases}
\sqrt[m]{\text{Conf}_{\text{pred}}(F_i)}, &\text{Conf}_{\text{pred}}(F_i) > \xi,\\
0, &{\text { Otherwise}},
\end{cases}
\end{align}
where $m$ is a positive integer (in the subsequent experiments, $m$ is set to 3), $\text{Conf}_{\text{pred}}\left(F_i\right)$ represents the prediction confidence of flying bird object $i$, and $\xi$ is the confidence threshold parameter (also known as the Training Schedule parameter). The Training Schedule method is expressed as follows,
\begin{align}\label{eq:xi_ep}
\xi = 
\begin{cases}
\xi_0, &\text{EP} < e_1,\\
\frac{\xi_0\left( e_2 - \text{EP} \right)}{e_2 - e_1}, &e_1 \leq \text{EP} < e_2,\\
0, &e_2 \leq \text{EP}.
\end{cases}
\end{align}
The relationship between the confidence threshold parameter $\xi$ and the training progress EP is shown in Figure \ref{xi_ep_fig} (in the subsequent experiment, $\xi_0$ is set to 0.8, $e_1$ and $e_2$ are set to 10\% and 90\% respectively), indicating that with the progress of training, the confidence threshold is gradually reduced to allow more samples to participate in the training.

\subsection{The CPL-BC strategy for the FBOD model training}

\begin{algorithm}
\caption{The training strategy of the FBOD model based on CPL-BC}\label{algo1}
\begin{algorithmic}[1]
\Require Flying birds dataset $S$, (FBOD) model $\mathcal{N}_\text{FBOD}$, the number of iterations $T$;
\Ensure The model weight parameters $\textbf{w}_f=\bf{\Theta}_f$, $\textbf{w}_g=\bf{\Theta}_g$.
\State Let $T=T_0 + T_1$;
\State Initialize the model weight $\textbf{w}_f$ with simple gaussian, initialize $t = 0$;
\While{$t \neq T_0 $}
        \State $t = t + 1$;
        \State Select a batch of images and corresponding labels from $S$ randomly;
        \State Input the images into $\mathcal{N}_\text{FBOD}$ with fixed weight $\textbf{w}_f$ to get outputs;
        \State Input the outputs and the labels into Eq. \eqref{eq:tl}, update $\textbf{w}_f$ by GD;
\EndWhile
\State Freeze $\textbf{w}_f = \bf{\Theta}_f^{'}$;
\State Initialize the model weight $\textbf{w}_g$ with simple gaussian, initialize $t = 0$;
\While{$t \neq T_0 $}
        \State $t = t + 1$;
        \State Select a batch of images and corresponding labels from $S$ randomly;
        \State Input the images into $\mathcal{N}_\text{FBOD}$ with fixed weight $\textbf{w}_g$ to get outputs;
        \State Input the outputs and the labels into Eq. \eqref{eq:tl}, update $\textbf{w}_g$ by GD;
\EndWhile
\State Freeze $\textbf{w}_g = \bf{\Theta}_g^{'}$;
\State Initialize the weights of model $f$ and $g$ with $\textbf{w}_f=\bf{\Theta}_f^{'}$ and $\textbf{w}_g=\bf{\Theta}_g^{'}$, the sample weights $\boldsymbol{\Vec{v}}_f=0$ and $\boldsymbol{\Vec{v}}_g=0$, parameter $\xi = \xi_0$, and $t = 0$;
\While{$t \neq T_1 $}
        \State $t = t + 1$;
        \State Select a batch of images and corresponding labels from $S$ randomly;
        \State Fix $\textbf{w}_f$, input the images into model $f$, and input the outputs model $f$ into Eq. \eqref{eq:mf_example} to get the sample weight $\boldsymbol{\Vec{v}}_g$;
        \State Fix $\textbf{w}_g$, input the images into model $g$, and input the outputs mode $g$ into Eq. \eqref{eq:mf_example} to get the sample weight $\boldsymbol{\Vec{v}}_f$;
        \State Input the outputs of model $f$, labels and $\boldsymbol{\Vec{v}}_f$ into Eq. \eqref{eq:wlf_fbod}, update $\textbf{w}_f$ by GD;
        \State Input the outputs of model $g$, labels and $\boldsymbol{\Vec{v}}_g$ into Eq. \eqref{eq:wlf_fbod}, update $\textbf{w}_g$ by GD;
        \State Update $\xi$ through Eq. \eqref{eq:xi_ep});// To include more hard samples
\EndWhile
\State Freeze $\textbf{w}_f=\bf{\Theta}_f$, $\textbf{w}_g=\bf{\Theta}_g$.
\end{algorithmic}
\end{algorithm}

Algorithm \ref{algo1} presents the pseudocode for applying the CPL-BC strategy to train the FBOD models. Initially, two FBOD models of identical structure are pre-trained using the entire dataset (model priors). During this pre-training phase, a conventional strategy is employed without differentiating sample difficulty or weighing losses. Training stops before overfitting to hard samples, typically after a relatively low number of iterations, to ensure the models initially can distinguish between easy and hard samples. Subsequently, the CPL-BC strategy is adopted for further training. This strategy aims to enable the models to progressively learn features of flying bird objects, starting with easy samples and moving to harder ones. In CPL-BC, the two models assess each other's sample difficulty based on prediction confidence, mitigating the issue of cumulative bias that may arise when a single model selects samples.

\section{Experiments}\label{Experiments}

To verify that the CPL-BC strategy proposed in this paper is effective and advanced in the training process of the FBOD model for surveillance video, we designed and carried out quantitative and qualitative comparative experiments. Specifically, we compare four other model training strategies: All Sample training strategy (AS, All Samples are used, and the model is trained by general training strategy that do not distinguish the difficulty of samples), Easy Sample training strategy (ES, Select Easy samples and train the model with general training strategy), loss-based Self-Paced Learning strategy \cite{kumar_SPL_for_latent_variable, jiang_easy_samples_first_sp_regularizers}, Self-Paced Learning strategy Based on Confidence (SPL-BC) \cite{spl_bc}. In the loss-based Self-Paced Learning strategy, four regularizers, Hard \cite{kumar_SPL_for_latent_variable}, Linear \cite{jiang_easy_samples_first_sp_regularizers}, Logarithmic \cite{jiang_easy_samples_first_sp_regularizers}, and Polynomial \cite{Gong_Polynomial_regularization} are used in comparison, which are respectively recorded as SPL-BH, SPL-BLine, SPL-BLog, and SPL-BPoly. For specific implementation details and parameter settings of all comparison training strategies, please refer to the relevant part of the literature \cite{spl_bc}.

Next, we will introduce the dataset used in the experiment (\ref{datestes}), the quantitative evaluation metrics (\ref{evaluation_metrics}), the details of the experiment's implementation (\ref{implementation_details}), and the contrastive analysis experiment in detail (\ref{comparative}).

 \subsection{Datasets}\label{datestes}

 To comprehensively evaluate the method proposed in this paper, we conducted experiments on datasets from two different scenarios. The first dataset is the Flying Bird object detection Dataset collected from Traction SubStation Surveillance Video (FBD-SV-TSS), which was used in several of our previous works \cite{fbod-bmi, fbod-sv, spl_bc_conference, spl_bc}. The second dataset is the published flying bird object detection dataset FBD-SV-2024 \cite{fbd-sv-2024}, which was collected from outdoor surveillance video.

FBD-SV-TSS: This dataset contains 120 videos with flying bird objects, totaling 28,353 images. Among them, 101 videos (24,898 images) are used as the training set, and 19 videos (3,455 images) are used as the test set. For more detailed information about this dataset, please refer to reference \cite{fbod-sv}.

FBD-SV-2024: This dataset contains 483 videos with flying bird objects, totaling 28,694 images. Among them, 400 videos (23,979 images) are used as the training set, and 83 videos (4,715 images) are used as the test set. For more detailed information about this dataset, please refer to reference \cite{fbd-sv-2024}.

 \subsection{Evaluation Metrics}\label{evaluation_metrics}

This paper evaluates the detection results of the model using the Average Precision (AP) metric from Pascal VOC 2007 \cite{2010_Pascal_VOC}, concerning the evaluation indicators of other object detection algorithms. Additionally, the False Detection Rate (FDR, the ratio of the number of falsely detected objects to the total number of detected objects) is also used to evaluate the model's performance.

 \subsection{Implementation Details}\label{implementation_details}

 The FBOD model previously designed in \cite{fbod-sv} will be utilized in this paper. Specifically, the input to the model is five consecutive 3-channel RGB images of size 672×384. At the same time, the output includes a confidence prediction feature map of size 336×192×1 and a bounding box regression feature map for flying bird objects of size 336×192×4. The output predicts the position of the flying bird object on the middle frame.
 
 All experiments are implemented under the PyTorch framework. The models are trained on an NVIDIA GeForce RTX 3090 GPU equipped with 24GB of VRAM. All experimental models are trained from scratch without using any pre-trained models. The trainable parameters of the network are randomly initialized using a normal distribution with a mean of 0 and a variance of 0.01. We chose Adam as the optimizer for the model, with an initial learning rate set to 0.001. During training, the learning rate decays by multiplying it by 0.95 after each epoch (i.e., one complete pass through all the training data), and the model undergoes 150 epochs. Among them, the first 50 iterations are the model's prior stage [All Samples are used, and the model is trained by general training methods first (All Samples Prior, ASP) Or select Easy Samples and train the model with general training methods first (Easy Samples Prior, ESP)], and the last 100 are the specific learning strategy stage. It is important to note that an iteration here refers to a complete traversal of the training set. In the particular learning strategy phase, all samples' predicted confidence or loss is calculated after each iteration. At the beginning of the next iteration, the saved prediction confidence or loss from the previous iteration is read and used to calculate the sample weight for the current iteration. During model training, the batch size is set to 8.

\subsection{Comparative Analysis Experiments}\label{comparative}

\subsubsection{The Auantitative Experiment}

The quantitative experimental results of the detection performance of the FBOD models trained with various training strategies on the FBD-SV-TSS dataset and tested on the test set are shown in Table \ref{tab:compare_FBD-SV-TSS}. The experimental results indicate that the FBOD model trained using the CPL-BC strategy (ESP) proposed in this paper achieves an $\text{AP}_{50}$ of 79.0\% on the FBD-SV-TSS dataset's test set. Specifically, this strategy improves the $\text{AP}_{50}$ by 2.8\% compared to the AS strategy, by 6.8\% compared to the ES strategy, and by 0.7\% compared to the SPL-BLine \cite{jiang_easy_samples_first_sp_regularizers} training strategy, which performs relatively best among the loss-based SPL strategies. Additionally, it improves by 0.8\% compared to the SPL-BC \cite{spl_bc} strategy. Furthermore, this strategy's False Detection Rate (FDR) is reduced by 6.5\% compared to the AS strategy but is slightly higher than that of the SPL-BC \cite{spl_bc} strategy. It is noteworthy that training the FBOD model with the CPL-BC strategy using a Prior based on Easy Samples (ESP) results in an $\text{AP}_{50}$ that is only 0.2\% higher than that of using a Prior based on the All Samples (ASP).
\begin{table}[!ht]
\caption{The results of various training strategies on FBD-SV-TSS dataset performance.\label{tab:compare_FBD-SV-TSS}}
\centering
\begin{tabular}{c| c| c c c c}
\hline
Training strategy & Prior type & $\text{AP}_{50}$ & $\text{AP}_{75}$  & AP & FDR\\
\hline
AS & - & 0.762 & 0.371 & 0.395 & 0.131\\
ES & - & 0.722 & 0.304 & 0.345 & 0.045\\
SPL-BH \cite{kumar_SPL_for_latent_variable} & ESP & 0.771 & 0.372 & 0.385 & 0.063\\
SPL-BLine \cite{jiang_easy_samples_first_sp_regularizers} & ESP & 0.783 & 0.346 & 0.389 & 0.056\\
SPL-BLog \cite{jiang_easy_samples_first_sp_regularizers} & ESP & 0.743 & 0.367 & 0.379 & 0.090\\
SPL-BPoly\cite{Gong_Polynomial_regularization}  & ESP & 0.780 & 0.366  & 0.390 & 0.078\\
SPL-BC \cite{spl_bc} & ESP & 0.782 & 0.369 & 0.398 & 0.053\\
CPL-BC (this paper) & ESP & 0.790 & 0.378 & 0.397 & 0.066\\
CPL-BC (this paper) & ASP & 0.788 & 0.370 & 0.396 & 0.055\\
\hline
\end{tabular}
\end{table}

Without differentiating the difficulty of samples and directly using all samples for model training, the model's training process is prone to interference from hard samples, resulting in a higher false detection rate for FBOD models trained with the AS strategy. Discarding the more hard samples and only using easy samples for model training can reduce the false detection rate. Still, the model's performance on hard samples will suffer, leading to a decline in overall detection performance. SPL strategies address this by allowing the model to start learning from easy samples and gradually transition to hard ones, progressively enhancing the model's ability to recognize objects and effectively mitigating the adverse effects of hard samples on model training. Therefore, FBOD models trained with SPL strategies typically exhibit high detection accuracy while maintaining a low false detection rate. The CPL-BC strategy proposed in this paper further overcomes the potential issue of cumulative sample selection bias that may arise during the training process of SPL strategies, further improving the detection performance of FBOD models. Additionally, it is worth noting that although the performance of FBOD models trained with the CPL-BC strategy using All Sample Prior is slightly lower than that of models using Easy Sample Prior, considering the cumbersome process of manually selecting Easy Samples, the strategy with All Sample Prior is also a viable option in practical applications.

The quantitative experimental results for the detection performance of the FBOD model trained on the FBD-SV-2024 dataset and evaluated on the test set using various training strategies are presented in Table \ref{tab:compare_FBD-SV-2024}. The experimental results demonstrate that the FBOD model trained using the CPL-BC strategy (ESP) proposed in this paper achieves an $\text{AP}_{50}$ of 73.2\% on the test set of the FBD-SV-2024 dataset. Specifically, this strategy improves the $\text{AP}_{50}$ by 1.3\% compared to the AS strategy and by 0.3\% compared to the SPL-BC \cite{spl_bc} strategy. Additionally, the FDR of this strategy is reduced by 2.7\% compared to the AS strategy and by 1\% compared to the SPL-BC strategy. The trends in various performance indicators are generally consistent with the previous results obtained on the FBD-SV-TSS dataset, and therefore, a detailed analysis is omitted here.
\begin{table}[!ht]
\caption{The results of various training strategies on FBD-SV-2024 dataset performance.\label{tab:compare_FBD-SV-2024}}
\centering
\begin{tabular}{c| c| c c c c}
\hline
Training strategy & Prior type & $\text{AP}_{50}$ & $\text{AP}_{75}$  & AP & FDR\\
\hline
AS & - & 0.719 & 0.341 & 0.371 & 0.150\\
ES & - & 0.682 & 0.310 & 0.340 & 0.108\\
SPL-BH \cite{kumar_SPL_for_latent_variable} & ESP & 0.711 & 0.329 & 0.363 & 0.114\\
SPL-BLine \cite{jiang_easy_samples_first_sp_regularizers} & ESP & 0.718 & 0.341 & 0.373 & 0.119\\
SPL-BLog \cite{jiang_easy_samples_first_sp_regularizers} & ESP & 0.713 & 0.349 & 0.372 & 0.108\\
SPL-BPoly\cite{Gong_Polynomial_regularization}  & ESP & 0.722 & 0.349  & 0.375 & 0.116\\
SPL-BC \cite{spl_bc} & ESP & 0.729 & 0.346 & 0.377 & 0.133\\
CPL-BC (this paper) & ESP & 0.732 & 0.355 & 0.377 & 0.123\\
CPL-BC (this paper) & ASP & 0.730 & 0.347 & 0.372 & 0.126\\
\hline
\end{tabular}
\end{table}

\subsubsection{The Qualitative Experiment}

\begin{figure*}[!ht]
\centering
\includegraphics[width=5.4in]{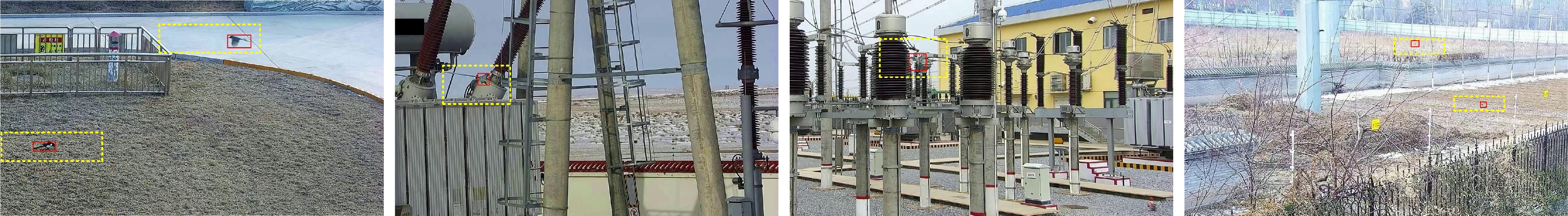}
\scriptsize{Situation 1~~~~~~~~~~~~~~~~~~~Situation 2~~~~~~~~~~~~~~~~~~~Situation 3~~~~~~~~~~~~~~~~~~~Situation 4}
\caption{Four typical Situations.}
\label{fig:Four_situations}
\end{figure*}
In the substation scenario, we selected four typical situations for a comparative qualitative experimental analysis of strategies such as AS, ES, SPL-BC, and CPL-BC. The images of these four situations are shown in Figure \ref{fig:Four_situations}. Specifically, Situation 1 contains two flying bird objects with relatively clear appearance features; Situations 2 and 3 each contain one flying bird object with prominent features but relatively complex backgrounds; and Situation 4 includes two flying bird objects that are not prominent and are relatively small in size within a single frame image.

\begin{figure*}[!htp]
    \centering
    \subfloat[Situation 1]{
       \rotatebox{90}{\scriptsize{~~~$3^\text{rd}$~~~~~$2^\text{nd}$~~~~~$1^\text{st}$}}
        \begin{minipage}[t]{0.465\linewidth}
        \centering
        \includegraphics[width=1\linewidth]{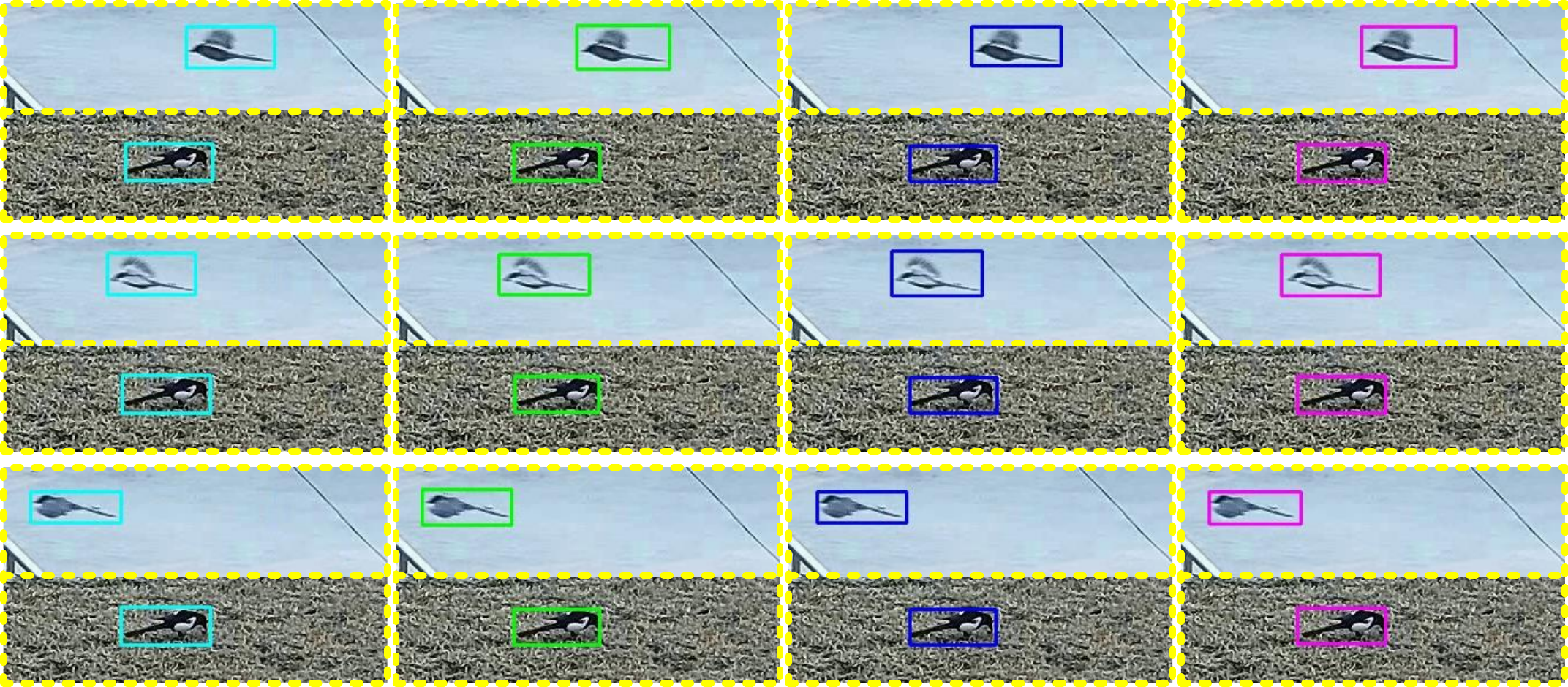}
        \label{s1_result}
        \scriptsize{~~~~AS~~~~~~~~~~~~ES~~~~~~~~~SPL-BC~~~~~CPL-BC}
        \end{minipage}
        }
    \subfloat[Situation 2]{
        \begin{minipage}[t]{0.465\linewidth}
        \centering
        \includegraphics[width=1\linewidth]{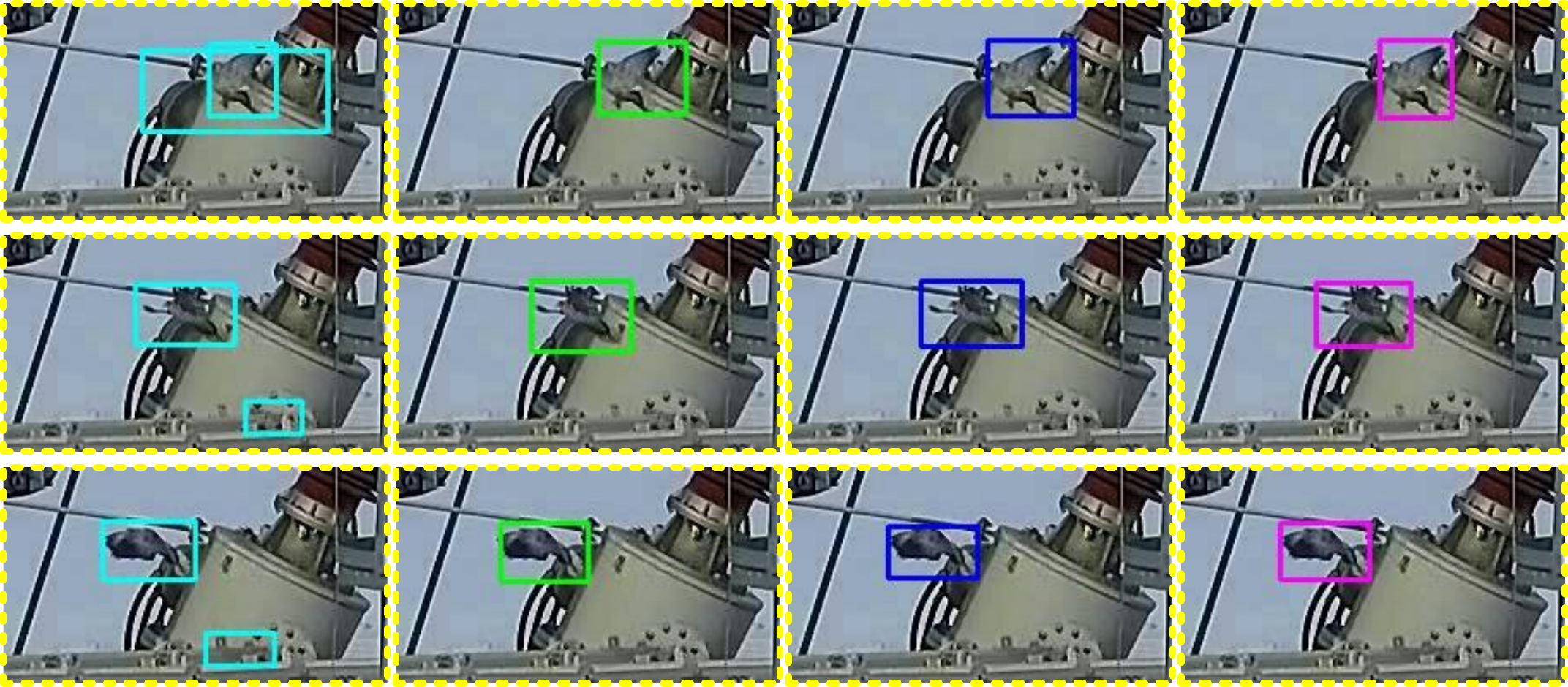}
        \label{s2_result}
        \scriptsize{~~~~AS~~~~~~~~~~~~ES~~~~~~~~~SPL-BC~~~~~CPL-BC}
        \end{minipage}
        }
    
    \vspace{-1mm}
    
    \subfloat[Situation 3]{
        \rotatebox{90}{\scriptsize{~~~$3^\text{rd}$~~~~~$2^\text{nd}$~~~~~$1^\text{st}$}}
        \begin{minipage}[t]{0.465\linewidth}
        \centering
        \includegraphics[width=1\linewidth]{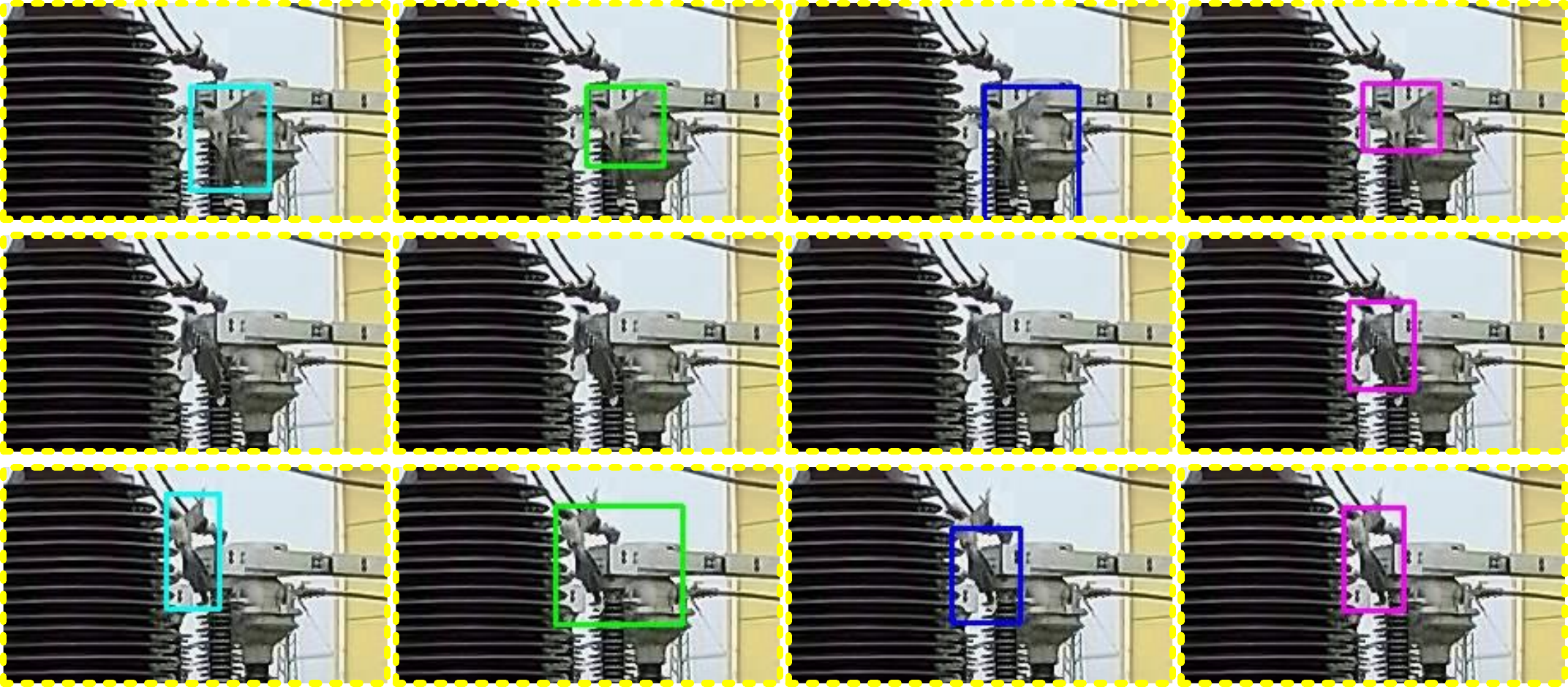}
        \label{s3_result}
        \scriptsize{~~~~AS~~~~~~~~~~~~ES~~~~~~~~~SPL-BC~~~~~CPL-BC}
        \end{minipage}
        }
    \subfloat[Situation 4]{
        \begin{minipage}[t]{0.465\linewidth}
        \centering
        \includegraphics[width=1\linewidth]{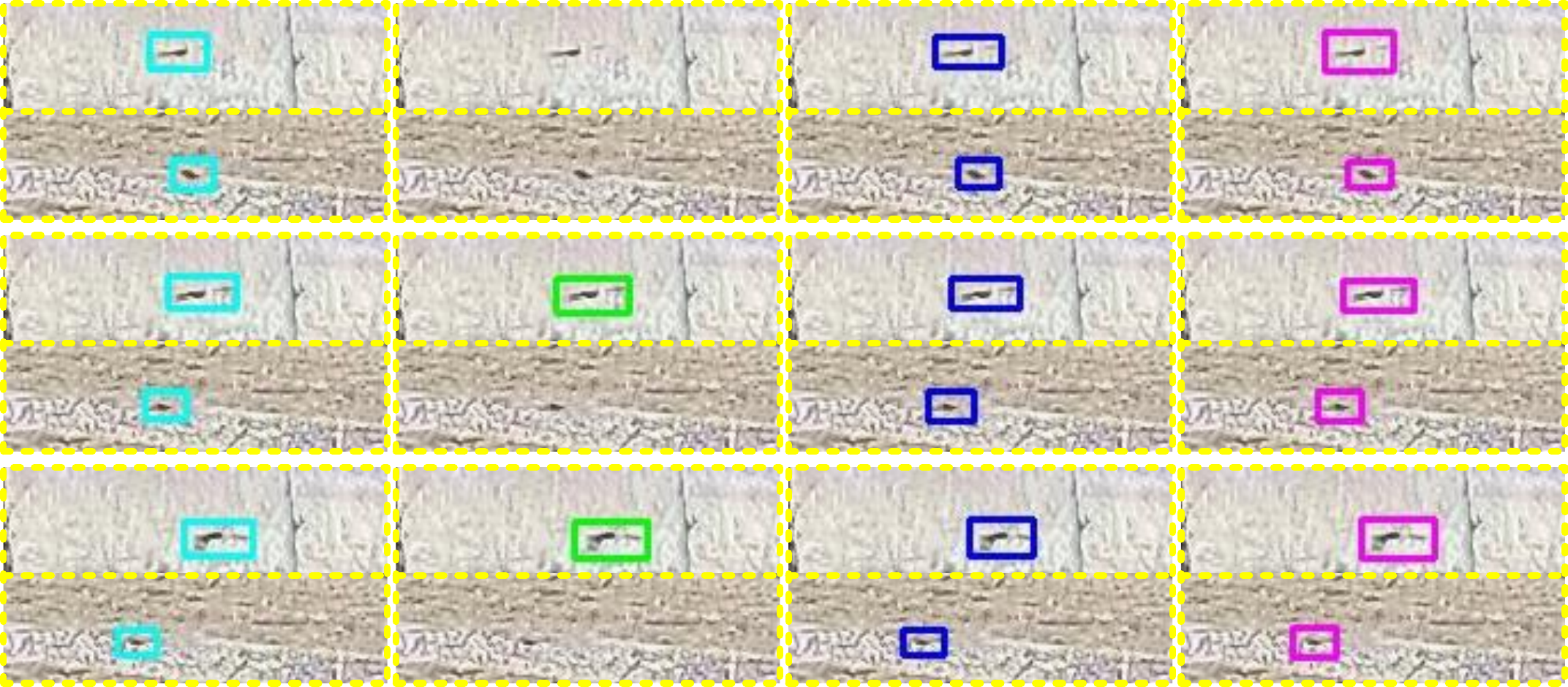}
        \label{s4_result}
        \scriptsize{~~~~AS~~~~~~~~~~~~ES~~~~~~~~~SPL-BC~~~~~CPL-BC}
        \end{minipage}
        }
    \caption{ Comparison results of qualitative experiments.}
    \label{result}
\end{figure*}

The qualitative experimental comparison results are shown in Figure \ref{result} (this figure displays enlarged images of three consecutive frames from the yellow dashed region in Figure \ref{fig:Four_situations}). In Situation 1 (Figure \ref{result}\subref{s1_result}), the FBOD models trained using all strategies successfully detected the flying bird objects. However, in Situation 2 (Figure \ref{result}\subref{s2_result}), although all models trained with different strategies detected the flying bird objects, the FBOD model trained with the AS strategy exhibited false detections. In Situation 3 (Figure \ref{result}\subref{s3_result}), except for the CPL-BC strategy, all other FBOD models trained with different strategies experienced missed detections. As for Situation 4 (Figure \ref{result}\subref{s4_result}), only the FBOD model trained with the ES strategy had missed detections.

The qualitative experimental results further validate the following points: Firstly, when training the FBOD model using the AS strategy, it is prone to false detections due to being easily influenced by hard samples. Secondly, when comparing the AS strategy with the ES strategy, we found that simply filtering out hard samples can address some false detection issues. Still, it more easily leads to missed detections. Furthermore, the SPL-BC \cite{spl_bc} strategy employs a gradual process, enabling the model to transition from easy samples to hard samples, thereby reducing false detections without increasing missed detections. Lastly, the CPL-BC strategy further optimizes the mechanism for selecting easy samples in the SPL-BC \cite{spl_bc} strategy, resulting in an even better detection performance.

\section{Conclusion}\label{Conclusion}

The paper introduces a novel model training strategy, termed Co-Paced Learning strategy Based on Confidence (CPL-BC). This strategy involves maintaining and training two models concurrently. These models independently select easy samples with prediction confidences exceeding a predetermined threshold for the other model to train on, thereby mitigating the accumulation of selection bias from a single model. As the training progresses, a specific strategy is employed to gradually reduce this threshold, enabling more samples to participate in the training process.

When applying the CPL-BC strategy to the training of the FBOD model, we devised a two-phase training approach. Initially, two FBOD models are pre-trained using a general training strategy to equip them with the capability to differentiate between easy and hard samples. Subsequently, the CPL-BC strategy is utilized to further train the FBOD models, enabling them to transition smoothly from easy to hard samples, thereby enhancing their ability to learn the characteristics of flying bird objects. Ultimately, the experimental results validate that the CPL-BC strategy can substantially improve the detection performance of the flying bird object detection models in surveillance videos.

However, it is important to note that although the models trained using the CPL-BC strategy exhibit superior detection performance compared to other strategies, the concurrent maintenance of two models during the implementation of the CPL-BC strategy results in a significantly slower training speed.

 \bibliographystyle{elsarticle-num} 
 \bibliography{cas-refs}

\begin{thebibliography}{10}
\expandafter\ifx\csname url\endcsname\relax
  \def\url#1{\texttt{#1}}\fi
\expandafter\ifx\csname urlprefix\endcsname\relax\def\urlprefix{URL }\fi
\expandafter\ifx\csname href\endcsname\relax
  \def\href#1#2{#2} \def\path#1{#1}\fi

\bibitem{a_new_skeleton_based_flying_bird_detection}
T.~WU, X.~LUO, Q.~XU, A new skeleton based flying bird detection method for low-altitude air traffic management, Chinese Journal of Aeronautics 31~(11) (2018) 2149--2164.

\bibitem{yolov5s_farm_bird_detction}
H.~Zhao, D.~Cai, Z.~Liang, Y.~Wang, An improved method for farm birds detection based on yolov5s, in: 2022 4th International Conference on Machine Learning, Big Data and Business Intelligence (MLBDBI), 2022, pp. 183--187.
\newblock \href {https://doi.org/10.1109/MLBDBI58171.2022.00043} {\path{doi:10.1109/MLBDBI58171.2022.00043}}.

\bibitem{automated_monitoring_for_birds}
C.~J. McClure, L.~Martinson, T.~D. Allison, Automated monitoring for birds in flight: Proof of concept with eagles at a wind power facility, Biological Conservation 224 (2018) 26--33.

\bibitem{2016_Hoffmann_multistatic_radar}
F.~Hoffmann, M.~Ritchie, F.~Fioranelli, A.~Charlish, H.~Griffiths, Micro-doppler based detection and tracking of uavs with multistatic radar, in: 2016 IEEE Radar Conference (RadarConf), 2016, pp. 1--6.
\newblock \href {https://doi.org/10.1109/RADAR.2016.7485236} {\path{doi:10.1109/RADAR.2016.7485236}}.

\bibitem{2017_Jahangirstaring_radar}
M.~Jahangir, C.~J. Baker, G.~A. Oswald, Doppler characteristics of micro-drones with l-band multibeam staring radar, in: 2017 IEEE Radar Conference (RadarConf), 2017, pp. 1052--1057.
\newblock \href {https://doi.org/10.1109/RADAR.2017.7944360} {\path{doi:10.1109/RADAR.2017.7944360}}.

\bibitem{object_survey}
Z.~Zou, K.~Chen, Z.~Shi, Y.~Guo, J.~Ye, Object detection in 20 years: A survey, Proceedings of the IEEE 111~(3) (2023) 257--276.
\newblock \href {https://doi.org/10.1109/JPROC.2023.3238524} {\path{doi:10.1109/JPROC.2023.3238524}}.

\bibitem{fbod-bmi}
Z.-W. Sun, Z.-X. Hua, H.-C. Li, H.-Y. Zhong, Flying bird object detection algorithm in surveillance video based on motion information, IEEE Transactions on Instrumentation and Measurement 73 (2024) 1--15.
\newblock \href {https://doi.org/10.1109/TIM.2023.3334348} {\path{doi:10.1109/TIM.2023.3334348}}.

\bibitem{fbod-sv}
Z.-W. Sun, Z.-X. Hua, H.-C. Li, Y.~Li, A flying bird object detection method for surveillance video, IEEE Transactions on Instrumentation and Measurement 73 (2024) 1--14.
\newblock \href {https://doi.org/10.1109/TIM.2024.3435183} {\path{doi:10.1109/TIM.2024.3435183}}.

\bibitem{spl_bc_conference}
Z.~Sun, Z.~Hua, H.~Li, \href{https://doi.org/10.1145/3672919.3673000}{A training strategy of flying bird object detection model based on improved self-paced learning algorithm}, in: Proceedings of the 2024 3rd International Conference on Cyber Security, Artificial Intelligence and Digital Economy, CSAIDE '24, Association for Computing Machinery, New York, NY, USA, 2024, p. 444–450.
\newblock \href {https://doi.org/10.1145/3672919.3673000} {\path{doi:10.1145/3672919.3673000}}.
\newline\urlprefix\url{https://doi.org/10.1145/3672919.3673000}

\bibitem{spl_bc}
Z.-W. Sun, Z.-X. hua, H.-C. Li, Y.~Li, Self-paced learning strategy with easy sample prior based on confidence for the flying bird object detection model training (2024).
\newblock \href {http://arxiv.org/abs/2412.06306} {\path{arXiv:2412.06306}}.

\bibitem{lu_mentornet}
L.~Jiang, Z.~Zhou, T.~Leung, L.-J. Li, L.~Fei-Fei, \href{https://api.semanticscholar.org/CorpusID:51876228}{Mentornet: Learning data-driven curriculum for very deep neural networks on corrupted labels}, in: International Conference on Machine Learning, Stockholm, Sweden, 2017.
\newline\urlprefix\url{https://api.semanticscholar.org/CorpusID:51876228}

\bibitem{malach_when_to_update}
E.~Malach, S.~Shalev-Shwartz, Decoupling ``when to update" from ``how to update", in: Proceedings of the 31st International Conference on Neural Information Processing Systems, NIPS'17, Curran Associates Inc., Red Hook, NY, USA, 2017, p. 961–971.

\bibitem{Han_co_teacher}
B.~Han, Q.~Yao, X.~Yu, G.~Niu, M.~Xu, W.~Hu, I.~W. Tsang, M.~Sugiyama, Co-teaching: robust training of deep neural networks with extremely noisy labels, in: Proceedings of the 32nd International Conference on Neural Information Processing Systems, NIPS'18, Curran Associates Inc., Red Hook, NY, USA, 2018, p. 8536–8546.

\bibitem{shen_learning_with_bad_training_data}
Y.~Shen, S.~Sanghavi, \href{https://proceedings.mlr.press/v97/shen19e.html}{Learning with bad training data via iterative trimmed loss minimization}, in: Proceedings of the 36th International Conference on Machine Learning, Vol.~97 of Proceedings of Machine Learning Research, PMLR, Long Beach, California, USA, 2019, pp. 5739--5748.
\newline\urlprefix\url{https://proceedings.mlr.press/v97/shen19e.html}

\bibitem{kumar_SPL_for_latent_variable}
M.~Kumar, B.~Packer, D.~Koller, Self-paced learning for latent variable models, in: Advances in Neural Information Processing Systems, 2010, pp. 1189--1197.

\bibitem{jiang_easy_samples_first_sp_regularizers}
L.~Jiang, D.~Meng, T.~Mitamura, A.~G. Hauptmann, \href{https://doi-org-s.era.lib.swjtu.edu.cn:443/10.1145/2647868.2654918}{Easy samples first: Self-paced reranking for zero-example multimedia search}, in: Proceedings of the 22nd ACM International Conference on Multimedia, MM '14, Association for Computing Machinery, New York, NY, USA, 2014, p. 547–556.
\newblock \href {https://doi.org/10.1145/2647868.2654918} {\path{doi:10.1145/2647868.2654918}}.
\newline\urlprefix\url{https://doi-org-s.era.lib.swjtu.edu.cn:443/10.1145/2647868.2654918}

\bibitem{spl_diversity}
L.~Jiang, D.~Meng, S.-I. Yu, Z.~Lan, S.~Shan, A.~Hauptmann, \href{https://api.semanticscholar.org/CorpusID:9686483}{Self-paced learning with diversity}, in: Neural Information Processing Systems, 2014.
\newline\urlprefix\url{https://api.semanticscholar.org/CorpusID:9686483}

\bibitem{spl_matrix_factor}
Q.~Zhao, D.~Meng, L.~Jiang, Q.~Xie, Z.~Xu, A.~G. Hauptmann, Self-paced learning for matrix factorization, in: Proceedings of the Twenty-Ninth AAAI Conference on Artificial Intelligence, AAAI'15, AAAI Press, 2015, p. 3196–3202.

\bibitem{spl_cl}
L.~Jiang, D.~Meng, Q.~Zhao, S.~Shan, A.~G. Hauptmann, Self-paced curriculum learning, in: Proceedings of the Twenty-Ninth AAAI Conference on Artificial Intelligence, AAAI'15, AAAI Press, 2015, p. 2694–2700.

\bibitem{evolution_spl}
M.~Gong, H.~Li, D.~Meng, Q.~Miao, J.~Liu, Decomposition-based evolutionary multiobjective optimization to self-paced learning, IEEE Transactions on Evolutionary Computation 23~(2) (2019) 288--302.
\newblock \href {https://doi.org/10.1109/TEVC.2018.2850769} {\path{doi:10.1109/TEVC.2018.2850769}}.

\bibitem{Han_co_teaching}
B.~Han, Q.~Yao, X.~Yu, G.~Niu, M.~Xu, W.~Hu, I.~W. Tsang, M.~Sugiyama, Co-teaching: robust training of deep neural networks with extremely noisy labels, in: Proceedings of the 32nd International Conference on Neural Information Processing Systems, NIPS'18, Curran Associates Inc., Red Hook, NY, USA, 2018, p. 8536–8546.

\bibitem{closer_look_deep_networks}
D.~Arpit, S.~Jastrzundefinedbski, N.~Ballas, D.~Krueger, E.~Bengio, M.~S. Kanwal, T.~Maharaj, A.~Fischer, A.~Courville, Y.~Bengio, S.~Lacoste-Julien, A closer look at memorization in deep networks, in: Proceedings of the 34th International Conference on Machine Learning - Volume 70, ICML'17, JMLR.org, 2017, p. 233–242.

\bibitem{Gong_Polynomial_regularization}
M.~Gong, H.~Li, D.~Meng, Q.~Miao, J.~Liu, Decomposition-based evolutionary multiobjective optimization to self-paced learning, IEEE Transactions on Evolutionary Computation 23~(2) (2019) 288--302.

\bibitem{yanbo_SPL_implicit_regularization}
Y.~Fan, R.~He, J.~Liang, B.-G. Hu, Self-paced learning: An implicit regularization perspective, in: AAAI Conference on Artificial Intelligence, 2016.

\bibitem{zheng_Enhancing_Geometric_Factors_in_Model_Learning_and_Inference}
Z.~Zheng, P.~Wang, D.~Ren, W.~Liu, R.~Ye, Q.~Hu, W.~Zuo, Enhancing geometric factors in model learning and inference for object detection and instance segmentation, IEEE Transactions on Cybernetics 52~(8) (2022) 8574--8586.
\newblock \href {https://doi.org/10.1109/TCYB.2021.3095305} {\path{doi:10.1109/TCYB.2021.3095305}}.

\bibitem{fbd-sv-2024}
Z.-W. Sun, Z.-X. Hua, H.-C. Li, Z.-P. Qi, X.~Li, Y.~Li, J.-C. Zhang, Fbd-sv-2024: Flying bird object detection dataset in surveillance video, Scientific Data 12~(1) (2025) 530.

\bibitem{2010_Pascal_VOC}
M.~Everingham, L.~Van~Gool, C.~K.~I. Williams, J.~Winn, A.~Zisserman, \href{https://doi.org/10.1007/s11263-009-0275-4}{The pascal visual object classes (voc) challenge}, International Journal of Computer Vision 88~(2) (2010) 303--338.
\newblock \href {https://doi.org/10.1007/s11263-009-0275-4} {\path{doi:10.1007/s11263-009-0275-4}}.
\newline\urlprefix\url{https://doi.org/10.1007/s11263-009-0275-4}

\end{thebibliography}





\end{document}